\documentclass[sigconf, nonacm]{acmart}
\AtBeginDocument{%
  \providecommand\BibTeX{{%
    \normalfont B\kern-0.5em{\scshape i\kern-0.25em b}\kern-0.8em\TeX}}}

\usepackage{graphicx}

\copyrightyear{2021} 
\acmYear{2021} 
\setcopyright{acmlicensed}\acmConference[ICMI '21 Companion]{Companion Publication of the 2021 International Conference on Multimodal Interaction}{October 18--22, 2021}{Montréal, QC, Canada}
\acmBooktitle{Companion Publication of the 2021 International Conference on Multimodal Interaction (ICMI '21 Companion), October 18--22, 2021, Montréal, QC, Canada}
\acmPrice{15.00}
\acmDOI{10.1145/3461615.3485424}
\acmISBN{978-1-4503-8471-1/21/10}

\begin{document}


\title[You made me feel this way]{“You made me feel this way”: Investigating Partners’ Influence in Predicting Emotions in Couples’ Conflict Interactions using Speech Data}

% \authors{}
% \emails{}
% \affiliations{
% \institution{}
% \city{} 
% \country{}
% }


% \author{}
% \affiliation{%
%   \institution{}
%   \city{}
%   \country{}}
% \email{}

% \authors{
% George Boateng
% Peter Hilpert 
% Guy Bodenmann
% Mona Neysary
% Tobias Kowatsch
% }
% \emails{
% gboateng@ethz.ch
% peter.hilpert@unil.ch
% guy.bodenmann@psychologie.uzh.ch 
% m.neysari@psychologie.uzh.ch
% tkowatsch@ethz.ch
% }
% \affiliations{
% \institution{
% ETHZ
% University of Lausanne
% University of Zurich
% University of Zurich
% ETHZ
% }
% \city{
% Zurich
% Lausanne
% Zurich
% Zurich
% Zurich} 

% \country{
% Switzerland
% Switzerland
% Switzerland
% Switzerland
% Switzerland}
% }

\author{George Boateng}
\affiliation{%
  \institution{ETH Zurich}
  \city{Zurich}
  \country{Switzerland}}
\email{gboateng@ethz.ch}

\author{Peter Hilpert}
\affiliation{%
  \institution{University of Lausanne}
  \city{Lausanne}
  \country{Switzerland}}
\email{peter.hilpert@unil.ch}

\author{Guy Bodenmann}
\affiliation{%
  \institution{University of Zurich}
  \city{Zurich}
  \country{Switzerland}}
\email{guy.bodenmann@psychologie.uzh.ch}

\author{Mona Neysari}
\affiliation{%
  \institution{University of Zurich}
  \city{Zurich}
  \country{Switzerland}}
\email{m.neysari@psychologie.uzh.ch}

\author{Tobias Kowatsch}
\email{tkowatsch@ethz.ch}
\affiliation{%
  \institution{ETH Z{\"u}rich}
  \city{Zurich}
  \country{Switzerland}
}
\affiliation{%
  \institution{University of St. Gallen}
  \city{St. Gallen}
  \country{Switzerland}
}

\renewcommand{\shortauthors}{Boateng, et al.}

\begin{abstract}
\end{abstract}




\begin{abstract}
How romantic partners interact with each other during a conflict influences how they feel at the end of the interaction and is predictive of whether the partners stay together in the long term. Hence understanding the emotions of each partner is important. Yet current approaches that are used include self-reports which are burdensome and hence limit the frequency of this data collection. Automatic emotion prediction could address this challenge. Insights from psychology research indicate that partners’ behaviors influence each other’s emotions in conflict interaction and hence, the behavior of both partners could be considered to better predict each partner’s emotion. However, it is yet to be investigated how doing so compares to only using each partner’s own behavior in terms of emotion prediction performance. In this work, we used BERT to extract linguistic features (i.e., what partners said) and openSMILE to extract paralinguistic features (i.e., how they said it) from a data set of 368 German-speaking Swiss couples (N = 736 individuals) who were videotaped during an 8-minutes conflict interaction in the laboratory. Based on those features, we trained machine learning models to predict if partners feel positive or negative after the conflict interaction.  Our results show that including the behavior of the other partner improves the prediction performance. Furthermore, for men, considering how their female partners spoke is most important and for women considering what their male partner said is most important in getting better prediction performance. This work is a step towards automatically recognizing each partners’ emotion based on the behavior of both, which would enable a better understanding of couples in research, therapy, and the real world.
\end{abstract}


\maketitle

\section{Introduction}
Understanding the emotions partners feel during and after conflict interactions is important because of its long-term effects on couples’ relationship quality and stability \cite{Gottman1994}. Happy couples, for example, experience more positive and less negative emotions during conflict interactions compared to unhappy couples \cite{Levenson1994}. The current study focuses on one fundamental aspect of the conflict mechanism — how the emotional experience \textit{within} each partner is influenced by the behavioral exchange \textit{between} partners. 

A crucial aspect of conflict interaction in couples is how the behavioral exchange makes each person feel during and after the interaction \cite{Ruef2007}. But although both partners experience the same interaction, they can feel very differently about it. For example, if we assume that partner A shows contempt and criticizes partner B, we can assume that partner A might feel angry or superior whereas partner B might feel hurt or humiliated. Thus, the experience can be very different for the partner who communicates something compared to the partner who perceives it \cite{Butler2011}. This differentiation allows us to reflect on another mechanism, namely that the emotions a person experiences are the results of two kinds of influences. Obviously, a person’s emotional experience is constantly influenced by a partner's behavior as a kind of co-regulating force, talk turn by talk turn. In addition, however, each person has the ability to regulate one’s own emotional response (e.g., cognitive appraisal, emotion regulation) \cite{Gross2014}, which then affects one's own subsequent behavioral response. Thus, what partners experience emotionally during and at the end of a conflict interaction is a reflection of the co-regulation and self-regulation processes \cite{Boker2006}. 

To better understand emotions in couples and their impact on relationships, often self-report assessments are used in which each partner is asked to provide a rating of their own emotions right after an interaction, or partners are asked to watch a video recording of the interaction and provide continuous ratings using a joystick, for example, \cite{Ruef2007, Hilpert2020}. Self-reports are burdensome to complete and may not be collected frequently. This means that the relationship between behavior and emotions cannot be studied often. Thus, an automatic emotion recognition system would allow scaling of couples research. 

Various works have used linguistic features (i.e., \textit{what} has been said) and paralinguistic features (i.e., \textit{how} it was said) to predict the emotions of each partner in couples interactions more broadly \cite{black2010, lee2010, black2013, lee2014, xia2015, li2016, chakravarthula2015, tseng2016, tseng2017, chakravarthula2018, tseng2018} and in conflict interactions in particular \cite{Chakravarthula2019, boateng2020a}.  Most of these works have used observer ratings (perceived emotions) as labels rather than self-reports (one’s actual emotions). Hence, the prediction task becomes that of recognizing external individuals’ perception of each partner’s emotion rather than each partner’s emotion per their own assessment. Though similar, the latter is more challenging than the former for a number of reasons. First, the rating might be biased and may not reflect their actual emotions over the period the rating is for (e.g., the past 5 minutes). Whereas for observer ratings, coders are generally trained over several weeks, it is done by more than one person and various approaches are employed to resolve ratings that are not in agreement and ensure the validity of the labels. Second, the self-reported emotion may not be reflected in that partner’s behavior in comparison to observer ratings which are purely based on behavioral observation. 

Despite these challenges, insights from psychology research could be leveraged to make the prediction task easier. Specifically, given that partners’ behaviors influence each other’s emotions in conflict interaction, the behavior of both partners could be considered to better predict each partner’s end-of-conversation emotion. However, it is yet to be investigated how doing so compares to using each partner’s own behavior only in terms of emotion prediction performance. In this work, we used a dataset collected from 368 couples who were recorded during an 8-minute conflict interaction, extracted linguistic and paralinguistic features, and used machine learning approaches to predict how each partner felt directly after the conflict interaction (self-reported emotion). We answer the following research questions (RQs)

\textbf{RQ1:} \textit{How well can the end-of-conversation emotion of each partner be predicted by their own behavior — a combination of linguistic and paralinguistic data? (self-regulation)}

\textbf{RQ2:} \textit{How does the prediction performance change when including the other partner’s behavior — (a) linguistic only, (b) paralinguistic only, and (c) combination of linguistic and paralinguistic data? (co-regulation) }

Our contributions are (1) an evaluation of how well a partner’s own linguistic and paralinguistic features predict one’s own end-of-conversation emotion (2) an investigation of how the prediction performance changes when including one’s partner’s features (linguistic, paralinguistic, and both) (3) the use of a unique dataset — spontaneous, real-life, speech data collected from German-speaking, Swiss couples (n=368 couples, N=736 participants), which is the largest ever such dataset used in the literature for automatic recognition of partners’ end-of-conversation emotion. The insights from our work would advance the use of methods to automatically recognize the emotions of each partner which could enable research and applications to better understand couples’ relationships in therapy and the real world. 

The rest of the paper is organized as follows:  In Section 2, we describe our data collection, preprocessing and feature extraction, in Section 3, we describe our experiments and evaluation, in Section 4 we present and discuss our results, in Section 5, we present limitations and future work, and we conclude in Section 6.

\section{Methodology}
% We extracted linguistic features using BERT, and paralinguistic features using openSMILE and performed multimodal and dyadic fusion. 

\subsection{Data Collection and Preprocessing}
This work used data from a larger dyadic interaction laboratory project conducted at the premises of the University of Zurich, Switzerland over 10 years with 368 heterosexual German-speaking, Swiss couples (N=736 participants; age 20-80) \cite{Kuster2015, uzh2020}. The inclusion criterion was to have been in the current relationship for at least 1 year. Couples had to choose one problematic topic for the conflict interaction from a list of common problems, and participants were then videotaped as they discussed the selected issue for 8 minutes. The data used in this work had one interaction from each couple and consequently, 368 8-minute interactions.

After each conversation, each partner provided self-reported responses to the Multidimensional Mood questionnaire \cite{steyer1997} of their emotions on four bipolar dimensions — namely “good mood versus bad mood,” “relaxed versus angry,” “happy versus sad” and “calm versus stressed” — with the scale: 1 — very much, 2 — much, 3 — a little, 4 — a little, 5 — much, 6 — very much. In this work, we sort to focus on predicting emotional valence (positive or negative) based on Russell’s circumplex model of emotions \cite{Russell1980}. Hence, we used an average of the “good mood versus bad mood” and “happy versus sad” scales which enables us to get a more valid score since several dimensions that measure similar constructs are combined. We did not use the other two scales because their polarity also could represent the arousal dimension of emotion (low vs high arousal). We then binarized the averaged values similar to prior works (e.g., \cite{boateng2020a, black2010}) such that values greater than or equal to 3.5 were negative (0) and the rest were positive (1). Binarization enables us to map the data into Russell’s circumplex model of emotions which has 4 quadrants for emotions, further enabling its real-world utility — easily being able to tell which group of emotions are being felt by each partner using binarized valence and arousal dimensions.

The speech data were manually annotated with the start and end of each speaker’s turn, along with pauses and noise. This was necessary in order to later be able to extract linguistic and paralinguistic features for each partner separately. In addition, speech content of both partners was manually transcribed for each partner separately and stored in 15-second chunks. Given that Swiss German is mostly spoken with different dialects across Switzerland, the spoken words were written as the corresponding German word equivalent.

Some couples requested their data to be removed and some data were missing due to technical problems in data collection. Of the original 368 couples that took part in the study, we could use 338 samples for females (46 negative labels) and 341 samples for males (32 negative labels). The distribution highlights a significant class imbalance that is characteristic of real-world datasets and partners’ emotions as seen in other similar works (e.g., \cite{boateng2020a}).

\subsection{Linguistic Features}
We extracted linguistic features from the transcripts of the whole 8-minute interaction using a pre-trained model — Sentence-BERT (SBERT) \cite{Reimers2019}. Sentence-BERT is a modification of the BERT architecture with siamese and triplet networks to compute sentence embeddings such that semantically similar sentences are close in vector space. Sentence-BERT has been shown to outperform the mean and CLS token outputs of regular BERT models for semantic similarity and sentiment classification tasks. Given that the text is in German, we used the German BERT model \cite{germanbert} as SBERT’s Transformer model and the mean pooling setting. The German BERT model was pre-trained using the German Wikipedia dump, the OpenLegalData dump, and German news articles. The extraction resulted in a  768-dimensional feature vector. 

\subsection{Paralinguistic Features}
We extracted acoustic features from the voice recordings. First, we used the speaker annotations to get the acoustic signal for each gender separately. Next, we used openSMILE  \cite{Eyben2015} to extract the 88 eGeMAPS acoustic features which have been shown to be a minimalist set of features for affective recognition tasks  \cite{Eyben2010}. The features are extracted in 25 ms sequences and then various functions (e.g., mean, median, range, etc.) are computed over the sequences resulting in 88 features for the whole 8-minute audio. The original audio was encoded with 2 channels. As a result, we extracted the features for each channel resulting in a 176-dimensional feature vector.

\subsection{Multimodal and Dyadic Feature Fusion}
Given that emotions are reflected in what and how people say things, we performed multimodal fusion (early fusion) by concatenating the linguistic and paralinguistic features resulting in a 944-dimensional feature. We did this separately for each partner. This feature vector was used as the baseline approach to answer research question (1).

Additionally, we fused features from both partners to answer research question (2). Specifically, for partner A, we concatenated their multimodal feature vector with the features of partner B and used it to predict partner A’s emotion label. This process was done for partner B as well. In order to investigate which behavioral data of the interacting partner was most important in the prediction of the emotions, we included the features in the following order (1) linguistic only, (2) paralinguistic only, and (3) multimodal fusion of both. 

Consequently, we had four feature sets: (1) Multimodal fusion (baseline — own features), (2)  Multimodal + Dyadic Fusion (with partner’s linguistic only), (2) Multimodal + Dyadic Fusion (with partner’s paralinguistic features only) (4) Multimodal + Dyadic Fusion (with partner’s combined linguistic and paralinguistic only). These were passed to machine learning models to answer the two research questions. 

\section{Experiments and Evaluation}
We run experiments using scikit-learn \cite{Pedregosa2011} the following machine learning models: support vector machine (SVM) algorithm with linear and radial basis function kernel, and random forests. We trained models to perform binary classification of each partner's self-reported positive and negative emotion using different feature sets. We used the four feature sets described in the previous section. To train and evaluate the models, we used nested K-fold cross-validation (CV). The nested procedure consisted of utilising an “inner” run of 5-fold CV for hyperparameter tuning, followed by an “outer” run of 10-fold CV which utilizes the best values for each hyperparameter found by the “inner” run. We prevented data from the same couple from being in both the train and test folds,  thereby evaluating the model’s performance on data from unseen couples. As the data was imbalanced, we used the metric balanced accuracy which is the unweighted average of the recall of each class and confusion matrices for evaluation. We used the “balanced” hyperparameter for all the models to mitigate the class imbalance while training. We compare to a random baseline of 50\% balanced accuracy. 

\begin{table}[t]
\caption{Prediction Results of the Best Models for the Fusion Approaches
}
\label{tab:results}
\begin{tabular}{|l|c|c|}
\hline
\textbf{Approach} &
  \multicolumn{2}{c|}{\textbf{\begin{tabular}[c]{@{}c@{}}Balanced \\ Accuracy (\%)\end{tabular}}} \\ \hline
                             & \textbf{Male} & \textbf{Female} \\ \hline
Multimodal fusion (baseline) & 49.8          & 59              \\ \hline
\begin{tabular}[c]{@{}l@{}}Multimodal + Dyadic Fusion (with partner’s \\ combined linguistic and paralinguistic only)\end{tabular} &
  52.3 &
  63.2 \\ \hline
\begin{tabular}[c]{@{}l@{}}Multimodal + Dyadic Fusion (with partner’s \\ linguistic features only)\end{tabular} &
  53.5 &
  \textbf{64.8} \\ \hline
\begin{tabular}[c]{@{}l@{}}Multimodal + Dyadic Fusion (with partner’s  \\ paralinguistic only)\end{tabular} &
  \textbf{56.1} &
  59.9 \\ \hline
\end{tabular}
\end{table}

\section{Results and Discussion}
Our results are shown in Table \ref{tab:results}. The baseline approach with multimodal fusion was not better than chance in predicting men’s  emotions at the end of the conflict interaction (49.8\%). This is unexpected as this indicates that men’s own behaviors during the interaction are not related to how they feel at the end of the interaction. This might be due to self-regulation processes or not showing much emotions during the interaction. This is clearly different from women as their emotions can be predicted by their own behavior  (59\%). Thus, it seems that women seem to express their emotions more clearly in their behavior. These results of poorer prediction performance for men compared to women is consistent with the results of \cite{boateng2020a}.

Including features from the interacting partner improved the results for both men (52.3\%) and women (63.2\%). These results are consistent with psychology research that the behavior of partners’ have an effect on each other’s emotions in conflict interaction \cite{Gottman2014, Butler2011}. This is a crucial finding because (i) previous research shows that the behavior of one person influences the behavior of the other person \cite{Gottman1994} (ii) as well as that the emotional changes of one person affect the emotions of the other \cite{Butler2011}. However, this is the first study showing that behavioral features assessed during the conflict interaction can be used to predict one partner’s emotion at the end of the conversation. In addition, the improvement in women’s emotion prediction at the end of interaction is greater when including their partner’s linguistic data (64.8\%) whereas there is hardly any difference when including partner’s paralinguistic features. This is a surprising finding as women generally pay more attention to paralinguistic cues \cite{Gottman1992}. Notably, the results are different for men. The prediction for men’s emotions slightly increases when including their partner’s linguistic features but the prediction improves substantially when including women’s paralinguistic features. Although we do not know which specific paralinguistic features are the main drivers for predicting the emotions, this finding is in line with prior findings — when women “nag”, men experience strong negative physiologic reactions and tend to withdraw \cite{Gottman1992, Gottman1994}. Future research is needed to investigate which aspects of one partner’s behavior exactly cause their own emotion prediction performance to decline. In addition, these results have implications for the kind of behavioral information to consider to best predict each partner’s end-of-conversation emotions.

\begin{figure}[t]
  \centering
  \includegraphics[width=0.6\linewidth]{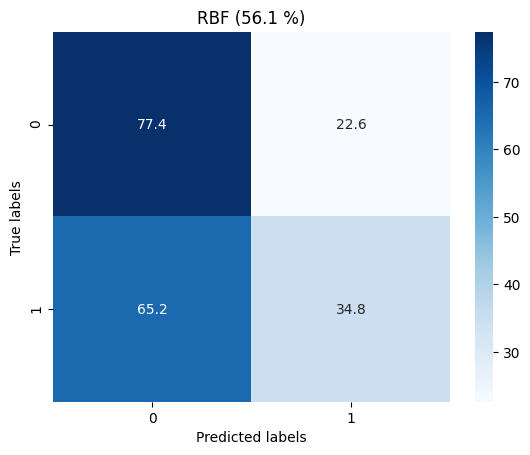}
  \caption{Confusion matrix for best performing model for male partners (Multimodal + Dyadic Fusion with partner’s linguistic features only)}
  \label{fig:conf_mat_male}
\end{figure}

\begin{figure}[t]
  \centering
  \includegraphics[width=0.6\linewidth]{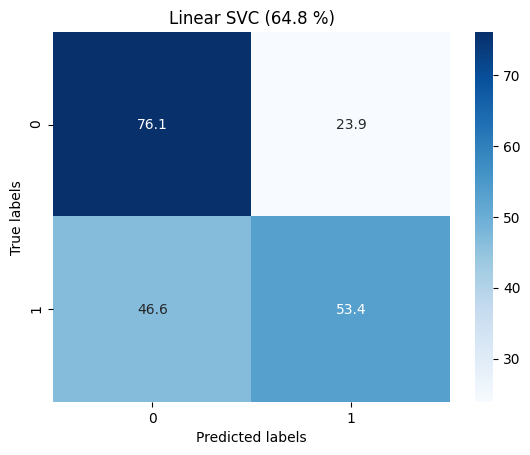}
  \caption{Confusion matrix for best performing model for female partners (Multimodal + Dyadic Fusion with partner’s paralinguistic features only) }
  \label{fig:conf_mat_female}
\end{figure}

We show the confusion matrices for the best models for the men and women in Figures \ref{fig:conf_mat_male} and \ref{fig:conf_mat_female} respectively. They reveal the models’ challenges at recognizing positive emotions (1), resulting in them being misclassified as negative emotions (0).

\section{Limitations and Future Work}
Further work is needed to investigate if these results are the same for couples in a different cultural context and also explore the effect of the interacting partners’ behavior at a more granular level such as on a talk-turn basis. More fine-grained emotion ratings may be needed to investigate that. Other fusion approaches like late fusion can be explored.

We used BERT as a feature extractor in this work. Generating domain-specific sentence embeddings via fine-tuning the BERT model and exploring deep transfer learning models for the paralinguistic features may improve the results. BERT models have been shown to encode gender and racial bias because of the models they are trained on. Further investigations are needed in the future of potential biases in prediction \cite{bender2021}.

Finally, we used manual annotations and transcripts. To accomplish true automatic emotion prediction, speaker annotations need to be done automatically and our approach needs to use and work for automated transcriptions. Current speech recognition systems do not work for this unique dataset given that couples speak Swiss German, which is (1) a spoken dialect and not written, and (2) varies across different parts of the German-speaking regions of Switzerland. Further work is needed to develop automatic speech recognition systems for Swiss German.

\section{Conclusion}
In this work, we investigated one’s own and the partner’s behavior in predicting end-of-conversation emotions in the context of conflict interactions in German-speaking Swiss couples. We extracted linguistic features using BERT and paralinguistic features using openSMILE. We fused both features in a multimodal approach for each partner. We also fused the features of both partners to predict the emotions of each partner. Our results show that including the behavior of the other partner improves the prediction performance. Furthermore, for men, considering \textit{how} their female partners spoke is most important, and for women considering \textit{what} their male partners said is most important in getting better prediction performance.  These insights have implications for the behavioral information to (not) include to better predict each partner’s end-of-conversation emotions which will enable a better understanding of couples relations in research, therapy, and the real world. 

\begin{acks}
Funding was provided by the Swiss National Science Foundation: CR12I1\_166348/1; CRSI11\_133004/1; P3P3P1\_174466; P300P1\_164582
\end{acks}

\balance{}

\bibliographystyle{ACM-Reference-Format}
\bibliography{refs}


\begin{thebibliography}{32}

%%% the Web address.

\ifx \showCODEN    \undefined \def \showCODEN     #1{\unskip}     \fi
\ifx \showDOI      \undefined \def \showDOI       #1{#1}\fi
\ifx \showISBNx    \undefined \def \showISBNx     #1{\unskip}     \fi
\ifx \showISBNxiii \undefined \def \showISBNxiii  #1{\unskip}     \fi
\ifx \showISSN     \undefined \def \showISSN      #1{\unskip}     \fi
\ifx \showLCCN     \undefined \def \showLCCN      #1{\unskip}     \fi
\ifx \shownote     \undefined \def \shownote      #1{#1}          \fi
\ifx \showarticletitle \undefined \def \showarticletitle #1{#1}   \fi
\ifx \showURL      \undefined \def \showURL       {\relax}        \fi
% invisible to TeX
\providecommand\bibfield[2]{#2}
\providecommand\bibinfo[2]{#2}
\providecommand\natexlab[1]{#1}
\providecommand\showeprint[2][]{arXiv:#2}

\bibitem[\protect\citeauthoryear{??}{ger}{[n.d.]}]%
        {germanbert}
 \bibinfo{year}{[n.d.]}\natexlab{}.
\newblock \bibinfo{title}{Open Sourcing German BERT}.
\newblock \bibinfo{howpublished}{\url{https://deepset.ai/german-bert}}.
\newblock
\newblock
\shownote{Accessed: 2020-05-1.}


\bibitem[\protect\citeauthoryear{Bender, Gebru, McMillan-Major, and
  Shmitchell}{Bender et~al\mbox{.}}{2021}]%
        {bender2021}
\bibfield{author}{\bibinfo{person}{Emily~M Bender}, \bibinfo{person}{Timnit
  Gebru}, \bibinfo{person}{Angelina McMillan-Major}, {and}
  \bibinfo{person}{Shmargaret Shmitchell}.} \bibinfo{year}{2021}\natexlab{}.
\newblock \showarticletitle{On the Dangers of Stochastic Parrots: Can Language
  Models Be Too Big?}. In \bibinfo{booktitle}{\emph{Proceedings of the 2021 ACM
  Conference on Fairness, Accountability, and Transparency}}.
  \bibinfo{pages}{610--623}.
\newblock


\bibitem[\protect\citeauthoryear{Black, Katsamanis, Lee, Lammert, Baucom,
  Christensen, Georgiou, and Narayanan}{Black et~al\mbox{.}}{2010}]%
        {black2010}
\bibfield{author}{\bibinfo{person}{Matthew Black}, \bibinfo{person}{Athanasios
  Katsamanis}, \bibinfo{person}{Chi-Chun Lee}, \bibinfo{person}{Adam~C
  Lammert}, \bibinfo{person}{Brian~R Baucom}, \bibinfo{person}{Andrew
  Christensen}, \bibinfo{person}{Panayiotis~G Georgiou}, {and}
  \bibinfo{person}{Shrikanth~S Narayanan}.} \bibinfo{year}{2010}\natexlab{}.
\newblock \showarticletitle{Automatic classification of married couples'
  behavior using audio features}. In \bibinfo{booktitle}{\emph{Eleventh annual
  conference of the international speech communication association}}.
\newblock


\bibitem[\protect\citeauthoryear{Black, Katsamanis, Baucom, Lee, Lammert,
  Christensen, Georgiou, and Narayanan}{Black et~al\mbox{.}}{2013}]%
        {black2013}
\bibfield{author}{\bibinfo{person}{Matthew~P Black},
  \bibinfo{person}{Athanasios Katsamanis}, \bibinfo{person}{Brian~R Baucom},
  \bibinfo{person}{Chi-Chun Lee}, \bibinfo{person}{Adam~C Lammert},
  \bibinfo{person}{Andrew Christensen}, \bibinfo{person}{Panayiotis~G
  Georgiou}, {and} \bibinfo{person}{Shrikanth~S Narayanan}.}
  \bibinfo{year}{2013}\natexlab{}.
\newblock \showarticletitle{Toward automating a human behavioral coding system
  for married couples’ interactions using speech acoustic features}.
\newblock \bibinfo{journal}{\emph{Speech communication}} \bibinfo{volume}{55},
  \bibinfo{number}{1} (\bibinfo{year}{2013}), \bibinfo{pages}{1--21}.
\newblock


\bibitem[\protect\citeauthoryear{Boateng, Sels, Kuppens, Hilpert, and
  Kowatsch}{Boateng et~al\mbox{.}}{2020}]%
        {boateng2020a}
\bibfield{author}{\bibinfo{person}{George Boateng}, \bibinfo{person}{Laura
  Sels}, \bibinfo{person}{Peter Kuppens}, \bibinfo{person}{Peter Hilpert},
  {and} \bibinfo{person}{Tobias Kowatsch}.} \bibinfo{year}{2020}\natexlab{}.
\newblock \showarticletitle{Speech Emotion Recognition among Couples using the
  Peak-End Rule and Transfer Learning}. In \bibinfo{booktitle}{\emph{Companion
  Publication of the 2020 International Conference on Multimodal Interaction
  (ICMI '20 Companion), October 25--29, 2020, Virtual event, Netherlands}}.
\newblock


\bibitem[\protect\citeauthoryear{Boker and Laurenceau}{Boker and
  Laurenceau}{2006}]%
        {Boker2006}
\bibfield{author}{\bibinfo{person}{Steven~M Boker} {and}
  \bibinfo{person}{Jean-Philippe Laurenceau}.} \bibinfo{year}{2006}\natexlab{}.
\newblock \showarticletitle{Dynamical systems modeling: An application to the
  regulation of intimacy and disclosure in marriage}.
\newblock \bibinfo{journal}{\emph{Models for intensive longitudinal data}}
  \bibinfo{volume}{63} (\bibinfo{year}{2006}), \bibinfo{pages}{195--218}.
\newblock


\bibitem[\protect\citeauthoryear{Butler}{Butler}{2011}]%
        {Butler2011}
\bibfield{author}{\bibinfo{person}{Emily~A Butler}.}
  \bibinfo{year}{2011}\natexlab{}.
\newblock \showarticletitle{Temporal interpersonal emotion systems: The
  “TIES” that form relationships}.
\newblock \bibinfo{journal}{\emph{Personality and Social Psychology Review}}
  \bibinfo{volume}{15}, \bibinfo{number}{4} (\bibinfo{year}{2011}),
  \bibinfo{pages}{367--393}.
\newblock


\bibitem[\protect\citeauthoryear{Chakravarthula, Baucom, and
  Georgiou}{Chakravarthula et~al\mbox{.}}{2018}]%
        {chakravarthula2018}
\bibfield{author}{\bibinfo{person}{Sandeep~Nallan Chakravarthula},
  \bibinfo{person}{Brian Baucom}, {and} \bibinfo{person}{Panayiotis Georgiou}.}
  \bibinfo{year}{2018}\natexlab{}.
\newblock \showarticletitle{Modeling Interpersonal Influence of Verbal Behavior
  in Couples Therapy Dyadic Interactions}.
\newblock \bibinfo{journal}{\emph{arXiv preprint arXiv:1805.09436}}
  (\bibinfo{year}{2018}).
\newblock


\bibitem[\protect\citeauthoryear{Chakravarthula, Gupta, Baucom, and
  Georgiou}{Chakravarthula et~al\mbox{.}}{2015}]%
        {chakravarthula2015}
\bibfield{author}{\bibinfo{person}{Sandeep~Nallan Chakravarthula},
  \bibinfo{person}{Rahul Gupta}, \bibinfo{person}{Brian Baucom}, {and}
  \bibinfo{person}{Panayiotis Georgiou}.} \bibinfo{year}{2015}\natexlab{}.
\newblock \showarticletitle{A language-based generative model framework for
  behavioral analysis of couples' therapy}. In \bibinfo{booktitle}{\emph{2015
  IEEE International Conference on Acoustics, Speech and Signal Processing
  (ICASSP)}}. IEEE, \bibinfo{pages}{2090--2094}.
\newblock


\bibitem[\protect\citeauthoryear{Chakravarthula, Li, Tseng, Reblin, and
  Georgiou}{Chakravarthula et~al\mbox{.}}{2019}]%
        {Chakravarthula2019}
\bibfield{author}{\bibinfo{person}{Sandeep~Nallan Chakravarthula},
  \bibinfo{person}{Haoqi Li}, \bibinfo{person}{Shao-Yen Tseng},
  \bibinfo{person}{Maija Reblin}, {and} \bibinfo{person}{Panayiotis Georgiou}.}
  \bibinfo{year}{2019}\natexlab{}.
\newblock \showarticletitle{Predicting Behavior in Cancer-Afflicted Patient and
  Spouse Interactions Using Speech and Language}.
\newblock \bibinfo{journal}{\emph{Proc. Interspeech 2019}}
  (\bibinfo{year}{2019}), \bibinfo{pages}{3073--3077}.
\newblock


\bibitem[\protect\citeauthoryear{Eyben, Scherer, Schuller, Sundberg, Andr{\'e},
  Busso, Devillers, Epps, Laukka, Narayanan, et~al\mbox{.}}{Eyben
  et~al\mbox{.}}{2015}]%
        {Eyben2015}
\bibfield{author}{\bibinfo{person}{Florian Eyben}, \bibinfo{person}{Klaus~R
  Scherer}, \bibinfo{person}{Bj{\"o}rn~W Schuller}, \bibinfo{person}{Johan
  Sundberg}, \bibinfo{person}{Elisabeth Andr{\'e}}, \bibinfo{person}{Carlos
  Busso}, \bibinfo{person}{Laurence~Y Devillers}, \bibinfo{person}{Julien
  Epps}, \bibinfo{person}{Petri Laukka}, \bibinfo{person}{Shrikanth~S
  Narayanan}, {et~al\mbox{.}}} \bibinfo{year}{2015}\natexlab{}.
\newblock \showarticletitle{The Geneva minimalistic acoustic parameter set
  (GeMAPS) for voice research and affective computing}.
\newblock \bibinfo{journal}{\emph{IEEE transactions on affective computing}}
  \bibinfo{volume}{7}, \bibinfo{number}{2} (\bibinfo{year}{2015}),
  \bibinfo{pages}{190--202}.
\newblock


\bibitem[\protect\citeauthoryear{Eyben, W{\"o}llmer, and Schuller}{Eyben
  et~al\mbox{.}}{2010}]%
        {Eyben2010}
\bibfield{author}{\bibinfo{person}{Florian Eyben}, \bibinfo{person}{Martin
  W{\"o}llmer}, {and} \bibinfo{person}{Bj{\"o}rn Schuller}.}
  \bibinfo{year}{2010}\natexlab{}.
\newblock \showarticletitle{Opensmile: the munich versatile and fast
  open-source audio feature extractor}. In
  \bibinfo{booktitle}{\emph{Proceedings of the 18th ACM international
  conference on Multimedia}}. \bibinfo{pages}{1459--1462}.
\newblock


\bibitem[\protect\citeauthoryear{Gottman}{Gottman}{1994}]%
        {Gottman1994}
\bibfield{author}{\bibinfo{person}{John~Mordechai Gottman}.}
  \bibinfo{year}{1994}\natexlab{}.
\newblock \bibinfo{booktitle}{\emph{What predicts divorce?: The relationship
  between marital processes and marital outcomes}}.
\newblock \bibinfo{publisher}{Lawrence Erlbaum Associates, Inc}.
\newblock


\bibitem[\protect\citeauthoryear{Gottman}{Gottman}{2014}]%
        {Gottman2014}
\bibfield{author}{\bibinfo{person}{John~Mordechai Gottman}.}
  \bibinfo{year}{2014}\natexlab{}.
\newblock \bibinfo{booktitle}{\emph{What predicts divorce?: The relationship
  between marital processes and marital outcomes}}.
\newblock \bibinfo{publisher}{Psychology Press}.
\newblock


\bibitem[\protect\citeauthoryear{Gottman and Levenson}{Gottman and
  Levenson}{1992}]%
        {Gottman1992}
\bibfield{author}{\bibinfo{person}{John~M Gottman} {and}
  \bibinfo{person}{Robert~W Levenson}.} \bibinfo{year}{1992}\natexlab{}.
\newblock \showarticletitle{Marital processes predictive of later dissolution:
  behavior, physiology, and health.}
\newblock \bibinfo{journal}{\emph{Journal of personality and social
  psychology}} \bibinfo{volume}{63}, \bibinfo{number}{2}
  (\bibinfo{year}{1992}), \bibinfo{pages}{221}.
\newblock


\bibitem[\protect\citeauthoryear{Gross}{Gross}{2014}]%
        {Gross2014}
\bibfield{author}{\bibinfo{person}{James~J Gross}.}
  \bibinfo{year}{2014}\natexlab{}.
\newblock \showarticletitle{Emotion regulation: Conceptual and empirical
  foundations.}
\newblock  (\bibinfo{year}{2014}).
\newblock


\bibitem[\protect\citeauthoryear{Hilpert, Brick, Fl{\"u}ckiger, Vowels,
  Ceulemans, Kuppens, and Sels}{Hilpert et~al\mbox{.}}{2020}]%
        {Hilpert2020}
\bibfield{author}{\bibinfo{person}{Peter Hilpert}, \bibinfo{person}{Timothy~R
  Brick}, \bibinfo{person}{Christoph Fl{\"u}ckiger}, \bibinfo{person}{Matthew~J
  Vowels}, \bibinfo{person}{Eva Ceulemans}, \bibinfo{person}{Peter Kuppens},
  {and} \bibinfo{person}{Laura Sels}.} \bibinfo{year}{2020}\natexlab{}.
\newblock \showarticletitle{What can be learned from couple research: Examining
  emotional co-regulation processes in face-to-face interactions.}
\newblock \bibinfo{journal}{\emph{Journal of Counseling Psychology}}
  \bibinfo{volume}{67}, \bibinfo{number}{4} (\bibinfo{year}{2020}),
  \bibinfo{pages}{475}.
\newblock


\bibitem[\protect\citeauthoryear{Kuster, Bernecker, Backes, Brandst{\"a}tter,
  Nussbeck, Bradbury, Martin, Sutter-Stickel, and Bodenmann}{Kuster
  et~al\mbox{.}}{2015}]%
        {Kuster2015}
\bibfield{author}{\bibinfo{person}{Monika Kuster}, \bibinfo{person}{Katharina
  Bernecker}, \bibinfo{person}{Sabine Backes}, \bibinfo{person}{Veronika
  Brandst{\"a}tter}, \bibinfo{person}{Fridtjof~W Nussbeck},
  \bibinfo{person}{Thomas~N Bradbury}, \bibinfo{person}{Mike Martin},
  \bibinfo{person}{Dorothee Sutter-Stickel}, {and} \bibinfo{person}{Guy
  Bodenmann}.} \bibinfo{year}{2015}\natexlab{}.
\newblock \showarticletitle{Avoidance orientation and the escalation of
  negative communication in intimate relationships.}
\newblock \bibinfo{journal}{\emph{Journal of Personality and Social
  Psychology}} \bibinfo{volume}{109}, \bibinfo{number}{2}
  (\bibinfo{year}{2015}), \bibinfo{pages}{262}.
\newblock


\bibitem[\protect\citeauthoryear{Lee, Black, Katsamanis, Lammert, Baucom,
  Christensen, Georgiou, and Narayanan}{Lee et~al\mbox{.}}{2010}]%
        {lee2010}
\bibfield{author}{\bibinfo{person}{Chi-Chun Lee}, \bibinfo{person}{Matthew
  Black}, \bibinfo{person}{Athanasios Katsamanis}, \bibinfo{person}{Adam~C
  Lammert}, \bibinfo{person}{Brian~R Baucom}, \bibinfo{person}{Andrew
  Christensen}, \bibinfo{person}{Panayiotis~G Georgiou}, {and}
  \bibinfo{person}{Shrikanth~S Narayanan}.} \bibinfo{year}{2010}\natexlab{}.
\newblock \showarticletitle{Quantification of prosodic entrainment in affective
  spontaneous spoken interactions of married couples}. In
  \bibinfo{booktitle}{\emph{Eleventh Annual Conference of the International
  Speech Communication Association}}.
\newblock


\bibitem[\protect\citeauthoryear{Lee, Katsamanis, Black, Baucom, Christensen,
  Georgiou, and Narayanan}{Lee et~al\mbox{.}}{2014}]%
        {lee2014}
\bibfield{author}{\bibinfo{person}{Chi-Chun Lee}, \bibinfo{person}{Athanasios
  Katsamanis}, \bibinfo{person}{Matthew~P Black}, \bibinfo{person}{Brian~R
  Baucom}, \bibinfo{person}{Andrew Christensen}, \bibinfo{person}{Panayiotis~G
  Georgiou}, {and} \bibinfo{person}{Shrikanth~S Narayanan}.}
  \bibinfo{year}{2014}\natexlab{}.
\newblock \showarticletitle{Computing vocal entrainment: A signal-derived
  PCA-based quantification scheme with application to affect analysis in
  married couple interactions}.
\newblock \bibinfo{journal}{\emph{Computer Speech \& Language}}
  \bibinfo{volume}{28}, \bibinfo{number}{2} (\bibinfo{year}{2014}),
  \bibinfo{pages}{518--539}.
\newblock


\bibitem[\protect\citeauthoryear{Levenson, Carstensen, and Gottman}{Levenson
  et~al\mbox{.}}{1994}]%
        {Levenson1994}
\bibfield{author}{\bibinfo{person}{Robert~W Levenson}, \bibinfo{person}{Laura~L
  Carstensen}, {and} \bibinfo{person}{John~M Gottman}.}
  \bibinfo{year}{1994}\natexlab{}.
\newblock \showarticletitle{Influence of age and gender on affect, physiology,
  and their interrelations: A study of long-term marriages.}
\newblock \bibinfo{journal}{\emph{Journal of personality and social
  psychology}} \bibinfo{volume}{67}, \bibinfo{number}{1}
  (\bibinfo{year}{1994}), \bibinfo{pages}{56}.
\newblock


\bibitem[\protect\citeauthoryear{Li, Baucom, and Georgiou}{Li
  et~al\mbox{.}}{2016}]%
        {li2016}
\bibfield{author}{\bibinfo{person}{Haoqi Li}, \bibinfo{person}{Brian Baucom},
  {and} \bibinfo{person}{Panayiotis Georgiou}.}
  \bibinfo{year}{2016}\natexlab{}.
\newblock \showarticletitle{Sparsely connected and disjointly trained deep
  neural networks for low resource behavioral annotation: Acoustic
  classification in couples' therapy}.
\newblock \bibinfo{journal}{\emph{arXiv preprint arXiv:1606.04518}}
  (\bibinfo{year}{2016}).
\newblock


\bibitem[\protect\citeauthoryear{Pedregosa, Varoquaux, Gramfort, Michel,
  Thirion, Grisel, Blondel, Prettenhofer, Weiss, Dubourg,
  et~al\mbox{.}}{Pedregosa et~al\mbox{.}}{2011}]%
        {Pedregosa2011}
\bibfield{author}{\bibinfo{person}{Fabian Pedregosa}, \bibinfo{person}{Ga{\"e}l
  Varoquaux}, \bibinfo{person}{Alexandre Gramfort}, \bibinfo{person}{Vincent
  Michel}, \bibinfo{person}{Bertrand Thirion}, \bibinfo{person}{Olivier
  Grisel}, \bibinfo{person}{Mathieu Blondel}, \bibinfo{person}{Peter
  Prettenhofer}, \bibinfo{person}{Ron Weiss}, \bibinfo{person}{Vincent
  Dubourg}, {et~al\mbox{.}}} \bibinfo{year}{2011}\natexlab{}.
\newblock \showarticletitle{Scikit-learn: Machine learning in Python}.
\newblock \bibinfo{journal}{\emph{the Journal of machine Learning research}}
  \bibinfo{volume}{12} (\bibinfo{year}{2011}), \bibinfo{pages}{2825--2830}.
\newblock


\bibitem[\protect\citeauthoryear{Reimers and Gurevych}{Reimers and
  Gurevych}{2019}]%
        {Reimers2019}
\bibfield{author}{\bibinfo{person}{Nils Reimers} {and} \bibinfo{person}{Iryna
  Gurevych}.} \bibinfo{year}{2019}\natexlab{}.
\newblock \showarticletitle{Sentence-bert: Sentence embeddings using siamese
  bert-networks}.
\newblock \bibinfo{journal}{\emph{arXiv preprint arXiv:1908.10084}}
  (\bibinfo{year}{2019}).
\newblock


\bibitem[\protect\citeauthoryear{Ruef and Levenson}{Ruef and Levenson}{2007}]%
        {Ruef2007}
\bibfield{author}{\bibinfo{person}{Anna~Marie Ruef} {and}
  \bibinfo{person}{Robert~W Levenson}.} \bibinfo{year}{2007}\natexlab{}.
\newblock \showarticletitle{Continuous measurement of emotion}.
\newblock \bibinfo{journal}{\emph{Handbook of emotion elicitation and
  assessment}} (\bibinfo{year}{2007}), \bibinfo{pages}{286--297}.
\newblock


\bibitem[\protect\citeauthoryear{Russell}{Russell}{1980}]%
        {Russell1980}
\bibfield{author}{\bibinfo{person}{James~A Russell}.}
  \bibinfo{year}{1980}\natexlab{}.
\newblock \showarticletitle{A circumplex model of affect.}
\newblock \bibinfo{journal}{\emph{Journal of personality and social
  psychology}} \bibinfo{volume}{39}, \bibinfo{number}{6}
  (\bibinfo{year}{1980}), \bibinfo{pages}{1161}.
\newblock


\bibitem[\protect\citeauthoryear{Steyer, Schwenkmezger, Notz, and Eid}{Steyer
  et~al\mbox{.}}{1997}]%
        {steyer1997}
\bibfield{author}{\bibinfo{person}{Rolf Steyer}, \bibinfo{person}{Peter
  Schwenkmezger}, \bibinfo{person}{Peter Notz}, {and} \bibinfo{person}{Michael
  Eid}.} \bibinfo{year}{1997}\natexlab{}.
\newblock \showarticletitle{Der Mehrdimensionale Befindlichkeitsfragebogen MDBF
  [Multidimensional mood questionnaire]}.
\newblock \bibinfo{journal}{\emph{G{\"o}ttingen, Germany: Hogrefe}}
  (\bibinfo{year}{1997}).
\newblock


\bibitem[\protect\citeauthoryear{Tseng, Baucom, and Georgiou}{Tseng
  et~al\mbox{.}}{2017}]%
        {tseng2017}
\bibfield{author}{\bibinfo{person}{Shao-Yen Tseng}, \bibinfo{person}{Brian~R
  Baucom}, {and} \bibinfo{person}{Panayiotis~G Georgiou}.}
  \bibinfo{year}{2017}\natexlab{}.
\newblock \showarticletitle{Approaching Human Performance in Behavior
  Estimation in Couples Therapy Using Deep Sentence Embeddings.}. In
  \bibinfo{booktitle}{\emph{INTERSPEECH}}. \bibinfo{pages}{3291--3295}.
\newblock


\bibitem[\protect\citeauthoryear{Tseng, Chakravarthula, Baucom, and
  Georgiou}{Tseng et~al\mbox{.}}{2016}]%
        {tseng2016}
\bibfield{author}{\bibinfo{person}{Shao-Yen Tseng},
  \bibinfo{person}{Sandeep~Nallan Chakravarthula}, \bibinfo{person}{Brian~R
  Baucom}, {and} \bibinfo{person}{Panayiotis~G Georgiou}.}
  \bibinfo{year}{2016}\natexlab{}.
\newblock \showarticletitle{Couples Behavior Modeling and Annotation Using
  Low-Resource LSTM Language Models.}. In
  \bibinfo{booktitle}{\emph{INTERSPEECH}}. \bibinfo{pages}{898--902}.
\newblock


\bibitem[\protect\citeauthoryear{Tseng, Li, Baucom, and Georgiou}{Tseng
  et~al\mbox{.}}{2018}]%
        {tseng2018}
\bibfield{author}{\bibinfo{person}{Shao-Yen Tseng}, \bibinfo{person}{Haoqi Li},
  \bibinfo{person}{Brian Baucom}, {and} \bibinfo{person}{Panayiotis Georgiou}.}
  \bibinfo{year}{2018}\natexlab{}.
\newblock \showarticletitle{" Honey, I Learned to Talk" Multimodal Fusion for
  Behavior Analysis}. In \bibinfo{booktitle}{\emph{Proceedings of the 20th ACM
  International Conference on Multimodal Interaction}}.
  \bibinfo{pages}{239--243}.
\newblock


\bibitem[\protect\citeauthoryear{University~of Zurich}{University~of
  Zurich}{[n.d.]}]%
        {uzh2020}
\bibfield{author}{\bibinfo{person}{UZH University~of Zurich}.}
  \bibinfo{year}{[n.d.]}\natexlab{}.
\newblock \bibinfo{title}{PASEZ Project-Impact of stress on relationship
  development of couples and children.}
\newblock
  \bibinfo{howpublished}{\url{http://www.dynage.uzh.ch/en/newsevents/news/news25.html}}.
\newblock
\newblock
\shownote{Accessed: 2021-05-1.}


\bibitem[\protect\citeauthoryear{Xia, Gibson, Xiao, Baucom, and Georgiou}{Xia
  et~al\mbox{.}}{2015}]%
        {xia2015}
\bibfield{author}{\bibinfo{person}{Wei Xia}, \bibinfo{person}{James Gibson},
  \bibinfo{person}{Bo Xiao}, \bibinfo{person}{Brian Baucom}, {and}
  \bibinfo{person}{Panayiotis~G Georgiou}.} \bibinfo{year}{2015}\natexlab{}.
\newblock \showarticletitle{A dynamic model for behavioral analysis of couple
  interactions using acoustic features}. In \bibinfo{booktitle}{\emph{Sixteenth
  Annual Conference of the International Speech Communication Association}}.
\newblock


\end{thebibliography}

\end{document}